\documentclass[journal]{IEEEtran}
\usepackage{tabu}
\usepackage[table]{xcolor}

\usepackage[T1]{fontenc}
\usepackage{babel}
\usepackage{array}
\usepackage{verbatim}
\usepackage{float}
\usepackage{multirow}
\usepackage{amsmath}
\usepackage{amssymb}
\usepackage{xcolor}
\usepackage{algorithm}
\usepackage{algorithmic}
\usepackage{graphicx}
\usepackage{xcolor}
\usepackage{booktabs}

\usepackage{physics}
\usepackage{graphicx}
\usepackage{float}
\usepackage{amsmath}
\usepackage{subfigure}
\usepackage{amsmath}
\usepackage{longtable}
\usepackage{booktabs}
\usepackage{float}
\usepackage{threeparttable}
\usepackage{xcolor}
\usepackage{diagbox}
\usepackage{cite}
\usepackage{hyperref}
\usepackage{color}
\usepackage{tabu}
\usepackage{amssymb}
\usepackage{xcolor}

\usepackage[table]{xcolor}
\usepackage{booktabs}
\usepackage{amsfonts}
\usepackage{amsmath}
\usepackage{amssymb}
\newcommand{\hm}[1]{{#1}}

\begin{document}
\title{Coarse-to-fine Task-driven Inpainting for Geoscience Images}

\author{Huiming Sun$^{\dagger}$, 
        Jin Ma$^{\dagger}$, 
        Qing Guo,  
        Qin Zou,  
        Shaoyue Song,   
        Yuewei Lin$^{*}$,  
        Hongkai Yu$^{*}$% <-this % stops a space
        
\thanks{Huiming Sun, Jin Ma, and Hongkai Yu are with Cleveland State University, Cleveland, OH, 44115, USA. Qing Guo is with Nanyang Technological University, Singapore. Qin Zou is with Wuhan University, Wuhan, China. Shaoyue Song is with Beijing University of Technology, Beijing, China. Yuewei Lin is with Brookhaven National Laboratory, Upton, NY, 11973, USA. This work was supported by NSF 2215388.}

\thanks{${\dagger}$ indicates the co-first authors. * Corresponding authors: Yuewei Lin (e-mail: ywlin@bnl.gov) and Hongkai Yu (e-mail: h.yu19@csuohio.edu).} 
% \thanks{Fundings TBD.}
}% <-this % stops a space


\maketitle 
\begin{abstract}
%The abstract goes here.
The processing and recognition of geoscience images have wide applications. \hm{Most of existing researches focus on understanding the high-quality  geoscience images by assuming that all the images are clear. However, in many real-world cases, the geoscience images might contain occlusions during the image acquisition. This problem actually implies the image inpainting problem in computer vision and multimedia. To the best of our knowledge, all the existing image inpainting algorithms learn to repair the occluded regions for a better visualization quality, they are excellent for natural images but not good enough for geoscience images by ignoring the geoscience related tasks. This paper aims to repair the occluded regions for a better geoscience task performance with the advanced visualization quality simultaneously, without changing the current deployed deep learning based geoscience models. Because of the complex context of geoscience images, we propose a coarse-to-fine encoder-decoder network with coarse-to-fine adversarial context discriminators to reconstruct the occluded image regions. Due to the limited data of geoscience images, we use a MaskMix based data augmentation method to exploit more information from limited geoscience image data. The experimental results on three public geoscience datasets for remote sensing scene recognition, cross-view geolocation and semantic segmentation tasks respectively show the effectiveness and accuracy of the proposed method.}
\end{abstract}


\IEEEpeerreviewmaketitle

%%%%%%%%% BODY TEXT
%-----------------------------------------------------------------------
%%%%%%%%% Introduction, By Dr. Yu
\section{Introduction}
\IEEEPARstart{T}{he} geoscience images have various representations, \textit{e.g.}, street-view  and aerial-view images. Geoscience image processing is an inter-disciplinary research with wide applications in computer vision and multimedia, such as remote sensing scene recognition~\cite{zou2015deep,song2019domain}, cross-view geolocation in urban environments~\cite{tian2017cross,wang2021each}, change detection~\cite{bruzzone2002adaptive,du2019unsupervised}, hyper-spectral classification~\cite{liu2021global,xie2020multiscale,sun2019low}, satellite-view object  detection~\cite{Xia_2018_CVPR,ding2019learning}, image captioning based remote understanding~\cite{li2020truncation,huang2020denoising},  semantic segmentation~\cite{bi2020polarimetric, liu2020afnet} and etc. 

Most of the existing researches in this area assume that all the obtained geoscience images are clear without occlusions. However, in many real-world cases, the geoscience images might contain occlusions during the image acquisition. For example, some regions of a street-view image might be occluded by a passing pedestrian or car close to the camera~\cite{pathak2016context}, and an aerial-view image by UAV (Unmanned Aerial Vehicle) might be partially occluded by a tall tower or a kite, and an aerial-view image by satellite might be occluded by thick cloud(s) to some extents~\cite{xie2017multilevel}. The occlusion challenge could result in the significant performance drop when using existing/deployed deep learning based geoscience models pre-trained on clean geoscience images. One straightforward way to relieve the challenge is to recover the occluded regions by using image inpainting methods. Then, the current deep geoscience models could be still functional without the need of any changes if we could well reconstruct the occluded image regions. 

\begin{figure} 
    \begin{centering}
        \includegraphics[width=1\columnwidth]{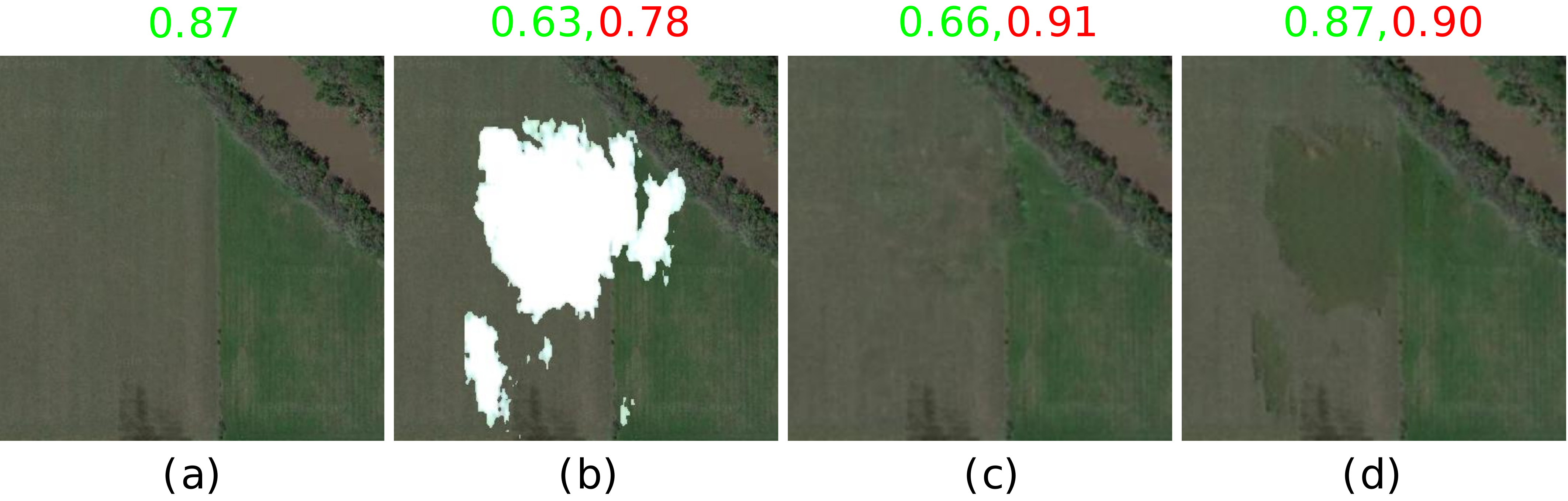}
    \par\end{centering}
    \caption{Illustration of the task-driven inpainting problem for geoscience images, taking the remote sensing scene recognition/classification task as an example: (a) a clean satellite image, (b) occluded image, (c) reconstruction by the image inpainting method CSA~\cite{Liu_2019_CSA}, (d) reconstruction by the proposed inpainting method. Green and red colored numbers indicate the  classification confidence of the correct class by a remote sensing scene recognition model (pre-trained on clean images) and the image quality SSIM of reconstruction respectively.}
    \label{fig:introduction}
\end{figure}

\begin{figure*}[htbp]
\begin{centering}
\includegraphics[width=0.9\textwidth]{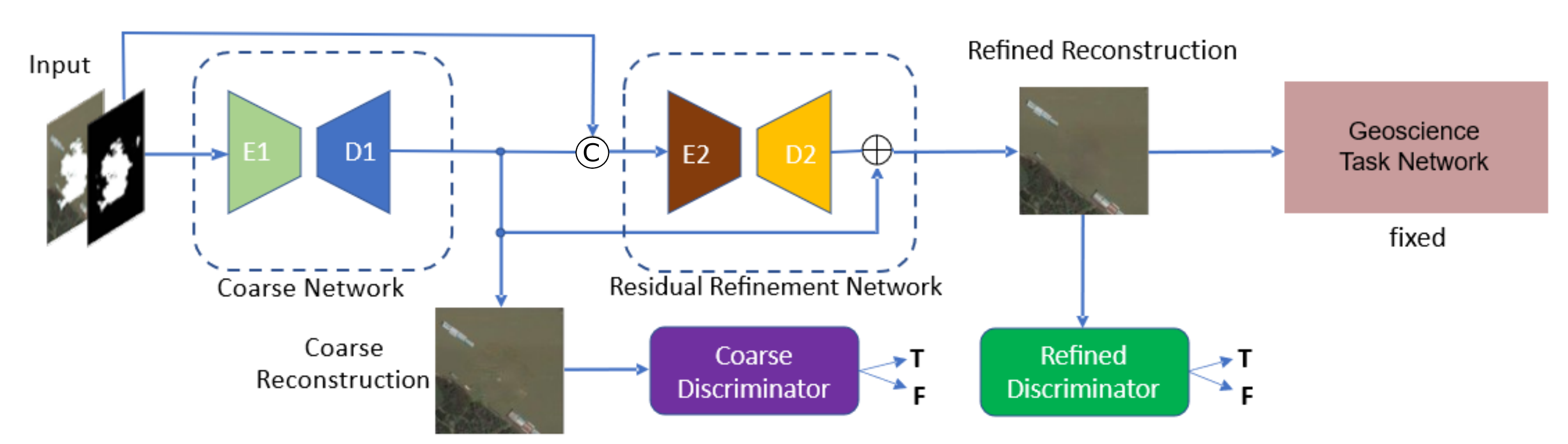}
\par\end{centering}
% to-do: add the Task Sub-network in figure and explain it in the end of Section 3.1 Network Overview before the Section 3.2 Loss functions. 
\caption{Overview of the proposed image inpainting network for geoscience images. From the Coarse Network to Residual Refinement Network with two adversarial context discriminators, we learn to reconstruct occluded regions in a coarse-to-fine way. \hm{Note that $\copyright$ is concatenation and $\oplus$ is element-wise summation.}}
\label{fig:architecture}	
\end{figure*}

As far as we know, the inpainting problem that focuses on geoscience images has not been systemically studied before. The existing image inpainting methods~\cite{pathak2016context,wang2018image,Liu_2019_CSA,wang2020structure,wu2021iid,xu2020e2i,jin2018patch} learn to recover the occluded regions for a better visualization quality, which are excellent for natural images but not good enough for geoscience images because they ignore the geoscience related tasks and fail to consider the domain speciality. For example, as shown in Fig.~\ref{fig:introduction}, the reconstruction visualization quality (SSIM) metric itself cannot fully represent the advanced geoscience task performance, since a higher SSIM 
score might not necessarily achieve better classification confidence in the remote sensing scene recognition/classification task. Therefore, we introduce the task-driven inpainting problem for geoscience images in this paper. 

\hm{There are several domain specialities for the task-driven image inpainting problem of geoscience images. First, the reconstruction objective is to largely improve the geoscience task performance with relatively high image quality, without changing the existing deep learning based geoscience task network pretrained on clean images. Second, the context of geoscience image is more complex than nature image without prior knowledge. Taking a face image as an example, if one eye is occluded, the deep learning based image inpainting model still knows the occluded region should be an eye there, because we have the prior knowledge of human face. Third, the dataset of geoscience images is typically much smaller than the regular nature images. For example, the Place2~\cite{zhou2017places} dataset that is frequently used for image inpainting has 10 million nature images. However, the geoscience images are relatively expensive to collect, so that many geoscience image datasets~\cite{zou2015deep,wang2016salient,li2020truncation} only have thousands of geoscience images.} 

We have made efforts in this paper to solve the challenges of  the above domain specialities. In this paper, we design a deep learning based image inpainting framework to embed the geoscience task network so that the reconstructed images could align with the geoscience task network. The geoscience task network can be replaced accordingly so as to fit different geoscience tasks, making the designed framework very flexible. Due to the above mentioned challenges, it might be difficult to simply learn a reliable deep learning based model within one stage to deal with the complex context of geoscience images, so we propose a coarse-to-fine encoder-decoder network to reconstruct the occluded regions with coarse-to-fine adversarial context discriminators. Due to the limited data of geoscience images, we design a MaskMix based data augmentation method to improve model robustness and overcome unforeseen corruptions by mixing different augmented  random occlusion masks during the model training. We expect that the inter-disciplinary research proposed in this paper could particularly benefit the geoscience image processing and recognition. In summary, the contributions of this paper are as the following. 

\begin{itemize}
    \item \hm{To the best of our knowledge, this paper is the first deep learning work for the task-driven inpainting problem of geoscience images. The reconstruction goal is to largely improve the geoscience task performance with relatively high image quality, without changing the existing pretrained deep geoscience task model. In addition, our method is general and flexible to be compatible with many different geoscience tasks.}    
    
    \item \hm{Due to the complex context in geoscience images, this paper proposes a deep coarse-to-fine encoder-decoder network to reconstruct the occluded image regions with coarse-to-fine adversarial context discriminators.} 
    
    \item \hm{Due to the limited training data in geoscience images, this paper proposes a MaskMix based data augmentation method to improve model robustness and overcome unforeseen corruptions.} 
    
\end{itemize}

In the following of this paper, Section~\ref{Sec:Related Work} reviews the related work. Section~\ref{Sec:Method} explains the proposed method. Experiment setting and  results are described in Section~\ref{Sec:Experiments}, followed by a conclusion in Section~\ref{Sec:Conclusions}.

%%%%%%%%% Related Work, final writing
\section{Related Work}\label{Sec:Related Work}

% Some methods restore the occluded cloud region of aerial-view satellite images with one more or a sequence of reference satellite images taken in different time~\cite{xu2016cloud,li2019cloud,zhang2020thick}. Taking extra reference images in different time requires higher data collection costs, while we would like to reconstruct the occluded region with just one geoscience image as the input. 

\subsection{Geoscience image processing}
Image processing and recognition have wide applications in geoscience and remote sensing. The Google Earth aerial-view images taken by satellites are desirable for remote sensing scene classification~\cite{zou2015deep,song2019domain}. The street-view Google Street images can be used to retrieve the aerial-view UAV or satellite images for the cross-view geolocation in urban  environments~\cite{tian2017cross,wang2021each}. Different objects are possible to be detected by some CNN based methods in the aerial-view Google Earth color images~\cite{Xia_2018_CVPR,ding2019learning}. Given an aerial-view Google Earth color image, image captioning can be utilized for remote sensing  understanding~\cite{li2020truncation,huang2020denoising}. Most of the existing researches assume that all the images are clear and high-quality. However, the geoscience images might contain occlusions during the image acquisition in many real-world cases, which result in significant difficulties for geoscience studies. For example, the street-view image might be partially blocked by a passing pedestrian or car close to the camera; the aerial-view image might be occluded by some thick clouds or flying objects. This paper focuses on the processing for the occluded geoscience images, \textit{i.e.}, either street-view or aerial-view color images. 

%   The image inpainting with regular holes means that the input 
 
\subsection{Image inpainting} 
The image inpainting problem aims to repair the occluded image  regions~\cite{pathak2016context,yu2018generative,liu2018image,jin2018patch,Li_2020_CVPR,Yi_2020_CVPR,wang2020vcnet,guo2021jpgnet,black2020evaluation,wang2020structure,wu2021iid,xu2020e2i}. Roughly, the image inpainting methods can be divided into two classes based on the input style: regular and irregular holes.  The image inpainting with regular holes means that the input is a rectangle-shape hole or multiple rectangle-shape holes, like the problems in~\cite{pathak2016context,yu2018generative}. The image inpainting with irregular holes means that the input shape is irregular, like the problems in~\cite{liu2018image,Li_2020_CVPR,Yi_2020_CVPR}. The context-aware information is very important to repair the occluded regions~\cite{pathak2016context} no matter for the regular or irregular input. This context-aware information could be enhanced in multiple ways, such as learning by the reconstruction and adversarial losses~\cite{pathak2016context}, the residual  aggregation~\cite{Yi_2020_CVPR}, contextual attention CNN layers~\cite{yu2018generative}, the partial   convolutions~\cite{liu2018image}, etc. All existing methods (\textit{e.g.},~\cite{wang2020structure,wu2021iid,xu2020e2i,jin2018patch}) are focused on the inpainting problem to improve the visualization quality for nature images, not for enhancing the geoscience related task performance, so they have ignored the domain speciality of geoscience images. 

\subsection{Inpainting for geoscience images}
% simplified conference version 
Existing inpainting methods for geoscience images can be divided into two classes. The first class belongs to non-deep learning based methods, which apply the traditional image processing based methods~\cite{cheng2013inpainting,shen2008map} to repair the contaminated region. Due to the advanced performance of deep learning, the second class leverages the CNN based methods for geoscience image inpainting~\cite{lin2018dense,dong2018inpainting, kuznetsov2020remote,lin2019remote,pan2020cloud} to remove the occlusions. 

% enriched journal version 
% Existing inpainting methods for geoscience images can be divided into two classes. The first class belongs to non-deep learning based methods, which apply the traditional image processing based methods, \textit{e.g.}, multi-channel nonlocal total variation~\cite{cheng2013inpainting,late2017remotely}, missing patch finding~\cite{fan2011pixel}, maximum a posteriori (MAP)-based algorithm~\cite{shen2008map} to repair the contaminated region. Due to the advanced performance of deep learning, the second class leverages the CNN based methods for geoscience image inpainting. For example, \cite{sidorov2019deep} uses the intrinsic properties of a CNN without any training for inpainting, and \cite{lin2018dense} performs  residual learning to learn the mappings from corrupted image to recovered image, and \cite{dong2018inpainting, kuznetsov2020remote} designs Generative Adversarial Networks (GAN) to recover remote sensing images with occlusions. Some researches design powerful CNN based networks~\cite{lin2019remote,pan2020cloud} to remove the occlusions of satellite images caused by thick clouds.  

The proposed task-driven inpainting problem for geoscience images has several differences compared to regular image inpainting on nature images: 1). The final goal is to improve the geoscience related task performance with the advanced visualization quality. 2). The context of geoscience image is more complex than nature image without prior knowledge. For example, toward the inpainting of face images, we can somewhat infer the corresponding occluded region because we have the prior knowledge of face. 3). The cost of collecting geoscience images is higher than that of collecting nature images, so the geoscience image dataset is typically much smaller than the regular nature image dataset. The existing geoscience image inpainting works are almost all focused on improving the repaired visualization quality, by ignoring the geoscience related task performance. In addition, the existing geoscience image inpainting methods ignore the above mentioned second and third special difficulties for geoscience images. Due to the complex contexts and limited training data, it might be challenging to learn a reliable deep learning model within one stage, so we propose to reconstruct the occluded region by a coarse-to-fine adversarial encoder-decoder structure and the assistance of a MaskMix based data augmentation during model training. 
%-------------------------------------------------------------------------
%%%%%%%%% Related Work, final writing

\section{Proposed Method}~\label{Sec:Method}
In this section, we explain the proposed network of task-driven image inpainting for geoscience images in details. The particular network structure is explained in Section~\ref{Network Overview}, while the loss functions for network training are presented in Section~\ref{Loss functions}, and the MaskMix based data augmentation is shown in Section~\ref{Sec:MaskMix}. 
 
\subsection{Network Overview}\label{Network Overview}
The overall network structure is shown in Fig.~\ref{fig:architecture}. It is an end-to-end deep learning based network. As we discussed above, reconstructing the occluded image regions is not easy due to the complex context and limited training data of geoscience images. It is hard to accomplish this task in one stage, so we learn to reconstruct the occluded image regions in a coarse-to-fine manner. In particular, it contains two Encoder-Decoder sub-networks: Coarse Network and Residual Refinement Network. \hm{The ``coarse-to-fine'' spirit is similar to some related computer vision  work~\cite{zhang2021automatic}.} 

% \subsubsection{Cloud Prediction Network} 
% We model the cloud prediction problem as an image segmentation problem to find the cloud region.  Given an input satellite image with clouds  I \in  \mathbb{R}^{H \times W \times 3}I \in  \mathbb{R}^{H \times W \times 3} with height HH pixels and width WW pixels, CPN learns a nonlinear mapping function \mathbf{f}\mathbf{f} to predict the binary segmentation mask M \in  \mathbb{R}^{H \times W \times 1}M \in  \mathbb{R}^{H \times W \times 1} for the cloud regions, where \mathbf{f}(I) \xrightarrow{} M \mathbf{f}(I) \xrightarrow{} M . In this task, there are many powerful CNN models for image segmentation. Therefore, we simply adopt the U-Net~\cite{ronneberger2015u} as the backbone for the CPN in this paper. 

\begin{figure}[htbp]
\centering
\includegraphics[width=1\columnwidth]{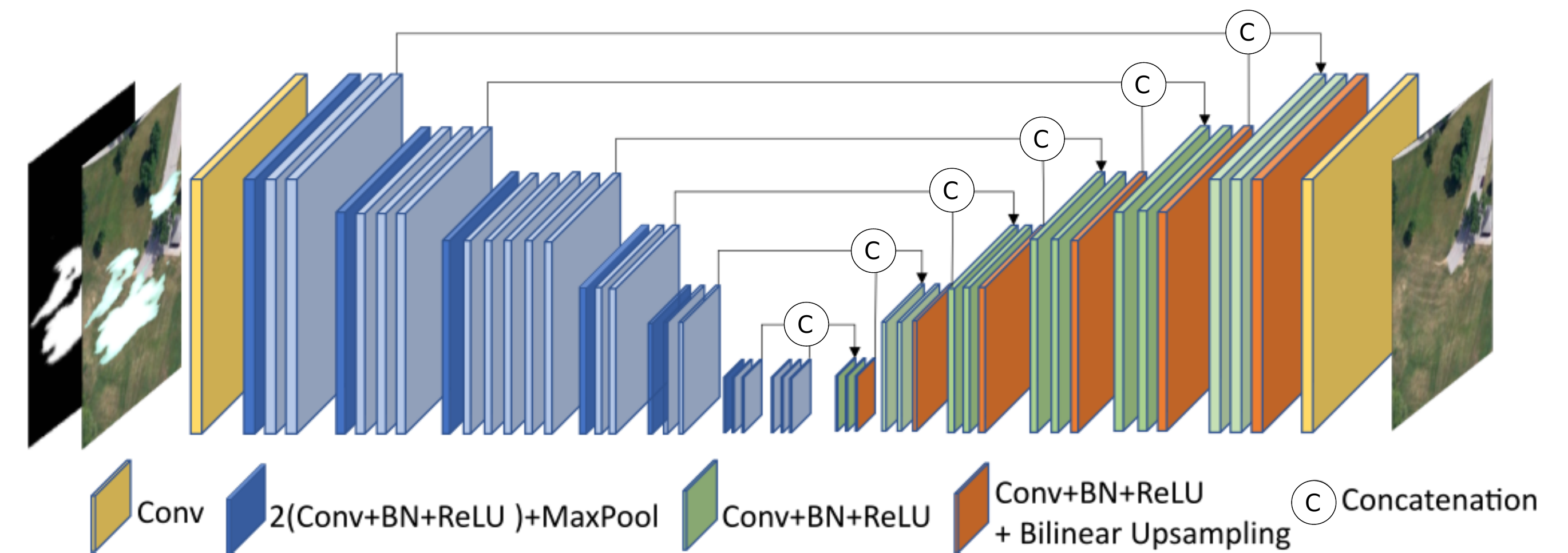}
\caption{Architecture of the Encoder-Decoder structure. \hm{It includes encoder, decoder and skip connections between the symmetric layers.}}
\label{fig:Generator}	
\end{figure}

\textbf{Encoder-Decoder Structure:} The Encoder-Decoder design has been widely used in different computer vision  tasks~\cite{badrinarayanan2017segnet}. Our proposed Coarse Network and Residual Refinement Network use the same Encoder-Decoder structure. Our backbone Encoder-Decoder structure is modified from~\cite{qin2019basnet}. The Encoder layers are from the ResNet-34~\cite{he2016deep}, and the Decoder is symmetric to the Encoder in network layers. Six skip connections are built between the symmetric layers of the Encoder and Decoder for a better deep feature fusion and perception. The Encoder implements the feature down-sampling and then the Decoder up-sample the features back to the same size of input image. The input and output of the encoder-decoder structure are $\mathbb{R}^{H \times W \times 4}$ and $ \mathbb{R}^{H \times W \times 3}$, respectively. The detailed organization of the used Encoder-Decoder structure is shown in Fig.~\ref{fig:Generator}. 

\textbf{Coarse Network:} The target of Coarse Network is to predict a coarse  reconstruction map $R_{coarse}$ directly from the input occluded image and the occlusion mask $M$. It follows the above mentioned encoder-decoder structure to reconstruct the occluded region.

\textbf{Residual Refinement Network:} \hm{Sometimes, it might be hard for a deep neural network to directly learn the prediction  well fitting the ground truth , which is because of the gradient vanishing problem in the backpropgation of the deep neural network. The ResNet work~\cite{he2016deep} shows that the residuals (by skip connection) could be learned to overcome the gradient vanishing problem in the backpropgation of the deep neural network.} The objective of Refinement Network is to predict a refined  reconstruction map $R_{refined}$ from the coarse reconstruction map $R_{coarse}$ given the occlusion mask $M$. Inspired  by~\cite{deng2018r3net,qin2019basnet}, the refinement network is described as a residual block that refines the coarse reconstruction map, then the problem is transferred to learn the residuals $E_{residual}$ between the reconstructed maps and the ground truth by the following equation:   

\begin{equation}
R_{refined} = R_{coarse} + E_{residual}, 
\label{eq:residual}
\end{equation}
where $R_{refined}=\mathbf{g} (R_{coarse}|M)$, and $\mathbf{g}$ indicates the nonlinear mapping function learned by the Refinement Network. In the network structure, we build a skip connect from the output of Coarse Network ($R_{coarse}$) to the end of the Refinement Network to implement this residual refinement.

\textbf{Coarse-to-fine Discriminators:} To enhance the context understanding of the occluded region, we deploy a Coarse Discriminator and a Refinement Discriminator for the Coarse Network and Refinement Network, respectively. The Coarse Discriminator is designed to distinguish the coarse reconstruction map $R_{coarse}$ as real or fake. Furthermore, the Refinement Discriminator is to judge the refined reconstruction map $R_{refined}$ as  real or fake. By introducing an adversarial learning between the encoder-decoder networks (generator) and the discriminators, our generator could produce the reconstructed maps close to the real ground-truth images so as to fool the discriminators in a coarse-to-fine way. Specifically, in order to make the discriminators  pay more attention on occluded regions, we process the reconstructed maps to change the input of discriminator, $D_{input}$, by the following equation:                
 
\begin{equation}
D_{input} = R \odot M + I \odot (1-M),  
 \label{eq:mask_dotproduct}
\end{equation}
where $R$ is a reconstructed map, $I$ is the input occluded image, $M$ is binary occlusion mask, and $\odot$ means pixel-level dot product. The processing in Eq.~(\ref{eq:mask_dotproduct}) are applied to both Coarse and Refine Discriminators with the corresponding reconstructed maps, respectively. The  network structure of each discriminator is adopted from the discriminator of the PatchGAN in the Pix2Pix model~\cite{isola2017image}, whose patch size is $70 \times 70$ for the real or fake classification. We also feed the input occluded image $I$ into the discriminators to simulate the conditional GAN. 

\textbf{Geoscience Task Network:} In order to accomplish the task-driven inpainting network, we incorporate the sub-network related to the specific geoscience task into the overall pipeline of the proposed method, as shown in Fig.~\ref{fig:architecture}. We treat the Geoscience Task Network as a fixed deep neural network that are pre-trained on clean images. This Geoscience Task Network is only used during training stage and discarded during testing stage. With such a design, the reconstructed image by our proposed inpainting method could be directly fed into the existing/deployed deep neural network for geoscience tasks during testing, so the existing deep learning based geoscience model will be compatible to both clean and occluded images without any changes. In other words, the proposed inpainting method can be used as a pre-processing procedure for the occluded geoscience images.      
%%%%%%%%%%%%%%%%%%%%%%%%%%%%%%%%%%%%%%%%%%%%%%%%%%%%%%%%%%%

\subsection{Loss Functions}\label{Loss functions}
In this section, we will introduce the detailed loss functions to train the proposed network.

\subsubsection{Reconstruction Loss}
Because the geoscience images have complex contexts with limited data, we design the comprehensive  reconstruction loss from multiple perspectives. We reconstruct the occluded region based on two considerations: pixel-level intensity  mismatching, and human perception difference. Based on these two considerations, we define the comprehensive  reconstruction loss based on two particular losses, $\mathcal{L}_1$ Loss, Perceptual Loss, to compare the reconstructed maps with the ground-truth image.   

\textbf{$\mathcal{L}_1$ Loss}: We expect to minimize the pixel-level intensity  mismatching between the reconstructed image $R$ and the ground-truth image $I_{gt}$, where the pixel-level intensity  mismatching is computed as their $L_1$ distance, as defined as: 

\begin{equation}
\mathcal{L}_{1} =  || R - I_{gt} ||_1.  
 \label{eq:L1-loss}
\end{equation}

Minimizing the $\mathcal{L}_1$ Loss makes $R$ and $I_{gt}$ have similar pixel-level intensity.

\textbf{Perceptual Loss}: We also want to minimize the human perception difference between $R$ and $I_{gt}$, which is computed by LPIPS (Learned Perceptual Image Patch Similarity) distance~\cite{zhang2018unreasonable} with a ImageNet pre-trained VGG16 model, as defined as: 

\begin{equation}
\mathcal{L}_{p} =  \mathbf{LPIPS}(R, I_{gt}),  
 \label{eq:LPIPS-loss}
\end{equation}
where $\mathbf{LPIPS}$ is a standard operation defined in~\cite{zhang2018unreasonable} to compute the $L_2$ distance of the activation feature maps in different CNN layers between two input images. The LPIPS metric is recently shown as more similar to the human perception seeing  an image. Minimizing the $\mathcal{L}_p$ Loss will force $R$ and $I_{gt}$ to have  consistent human perception response.

For the Coarse Network, we only use the $\mathcal{L}_1$ loss for its  reconstruction, while we use the summation of $\mathcal{L}_1$  Loss and Perceptual Loss for the Refinement Network's reconstruction.  

\subsubsection{Adversarial Loss}
Let us treat the proposed network in  Fig.~\ref{fig:architecture} except the Coarse Discriminator $D_c$ and Refinement Discriminator $D_r$ as a  coarse-to-fine generator $G$. We learn the generator $G$ and the discriminators $D_c$ and $D_r$ by the following adversarial loss:   

\begin{equation}\label{eq:GAN loss} 
\begin{split}
\min_{G} \max_{D_c, D_r} \mathcal{L}_{GAN}=  \mathbb{E}_{I, I_{gt}} [\log D_c(I,I_{gt}))] + \\ 
 \mathbb{E}_{I, G(I)} [\log (1-D_c(I, G(I)_c))] + \\
 \mathbb{E}_{I, I_{gt}} [\log D_r(I,I_{gt}))]  + \\
 \mathbb{E}_{I, G(I)} [\log (1-D_r(I, G(I)_r))], 
\end{split}
\end{equation}
where $I$ is the input occluded image, $I_{gt}$ is the ground-truth reconstructed image for $I$, $G(I)_c$ is the predicted coarse reconstruction map, $R_{coarse}$, and  $G(I)_r$ is the predicted refined reconstruction map, $R_{refined}$. $G$ tries to minimize the loss $\mathcal{L}_{GAN}$ against two adversarial $D_c$ and $D_r$ that would like to maximize it. The coarse and refinement discriminators try to classify the reconstructed region as real or fake in a coarse-to-fine manner. Since the context of geoscience images is complex with limited training data, we use the coarse-to-fine generator $G$ to reconstruct the occluded regions, and simultaneously we use the coarse-to-fine discriminators to adversarially improve the context reconstruction ability of the generator $G$.    

\subsubsection{Geoscience Task Loss}\label{task loss}
% Example, task1: RS scene classification loss, task2: * loss  
The geoscience task loss $\mathcal{L}_{T}$ is related to the specific geoscience task. \hm{In this paper, we applied the proposed network to three geoscience tasks as example:  remote sensing scene recognition, cross-view geolocation and semantic segmentation.} For example, in the task of remote sensing scene recognition, the internal purpose is image classification, so we could choose the image classification network like fixed VGG16~\cite{simonyan2014very} pretrained on clean images as the geoscience task network, where the corresponding cross entropy loss for image classification is used as the geoscience task loss. When applying our proposed method for cross-view geolocation, the geoscience task network could be the fixed LPN~\cite{wang2021each} pretrained on clean images, \hm{and the corresponding geoscience task loss is the summation of cross entropy losses over all image parts as defined in~\cite{wang2021each}. For the semantic segmentation task, we use 
HRNet~\cite{wang2020deep} pretrained on clean images as the geoscience task network and set the pixel-level cross-entropy loss as the geoscience task loss.} Minimizing the geoscience task loss $\mathcal{L}_{T}$ will make the reconstructed image well fit the fixed geoscience task network pretrained on clean images.

The overall loss function for training our proposed network is shown below as 

\begin{equation}
\begin{split}
    \mathcal{L}_{overall} & =  \mathcal{L}_{1} + \mathcal{L}_{P} + \mathcal{L}_{GAN} +  \lambda \mathcal{L}_{T}
%    \mathcal{L}_{T} & = \text{one of }  \{\mathcal{L}_{c}, \mathcal{L}_{g},\mathcal{L}_{s}\},
 \label{eq:L_overall}
\end{split}
\end{equation}
where $\lambda$ is the weight to balance the image quality of reconstruction and the geoscience task performance, \hm{ $\mathcal{L}_{T}$ depends on the specific geoscience task.} 

\begin{figure}[htbp]
\begin{centering}
\includegraphics[width=0.9\columnwidth]{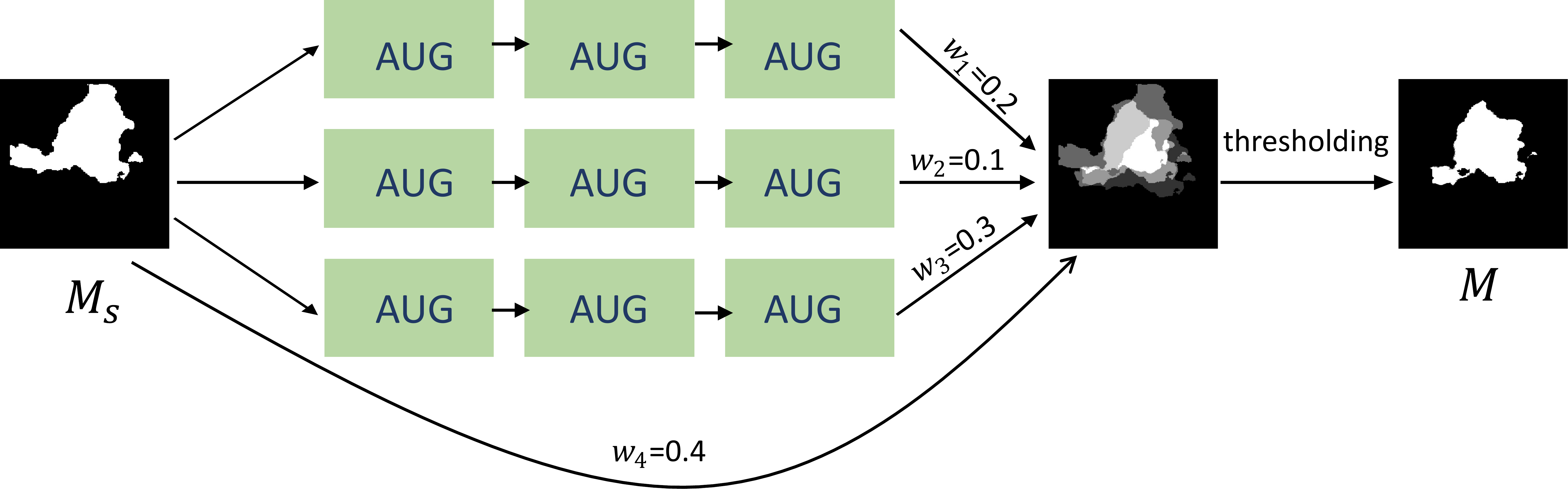}
\par\end{centering}
\caption{Illustration of MaskMix based data augmentation. AUG  indicates the random augmentation operation of ``translate", ``shear" and ``rotate".}
\label{fig:MaskMix}	
\end{figure} 

\begin{figure*}[htbp]
\begin{centering}
\includegraphics[width=0.95\textwidth]{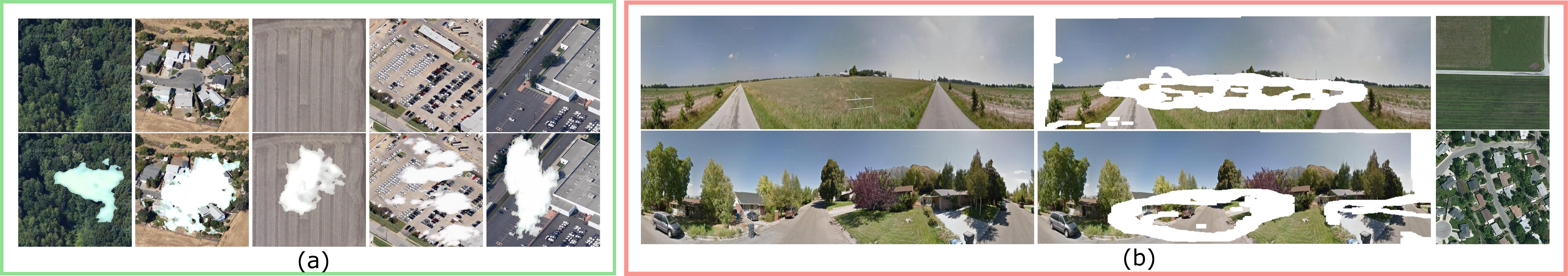}
\par\end{centering}
\caption{Datasets of image inpainting for two geoscience tasks: Remote Sensing (RS) scene recognition and cross-view geolocation: (a) RSSCN7~\cite{zou2015deep} dataset for RS scene recognition (Top: satellite view, Bottom: occluded satellite view), (b) CVUSA~\cite{zhai2017predicting} dataset for cross-view geolocation (From left to right: ground view, occluded ground view, corresponding satellite view of same location).} 
\label{fig:dataset}	
\end{figure*}

\begin{figure}[htbp]
\begin{centering}
\includegraphics[width=1\columnwidth]{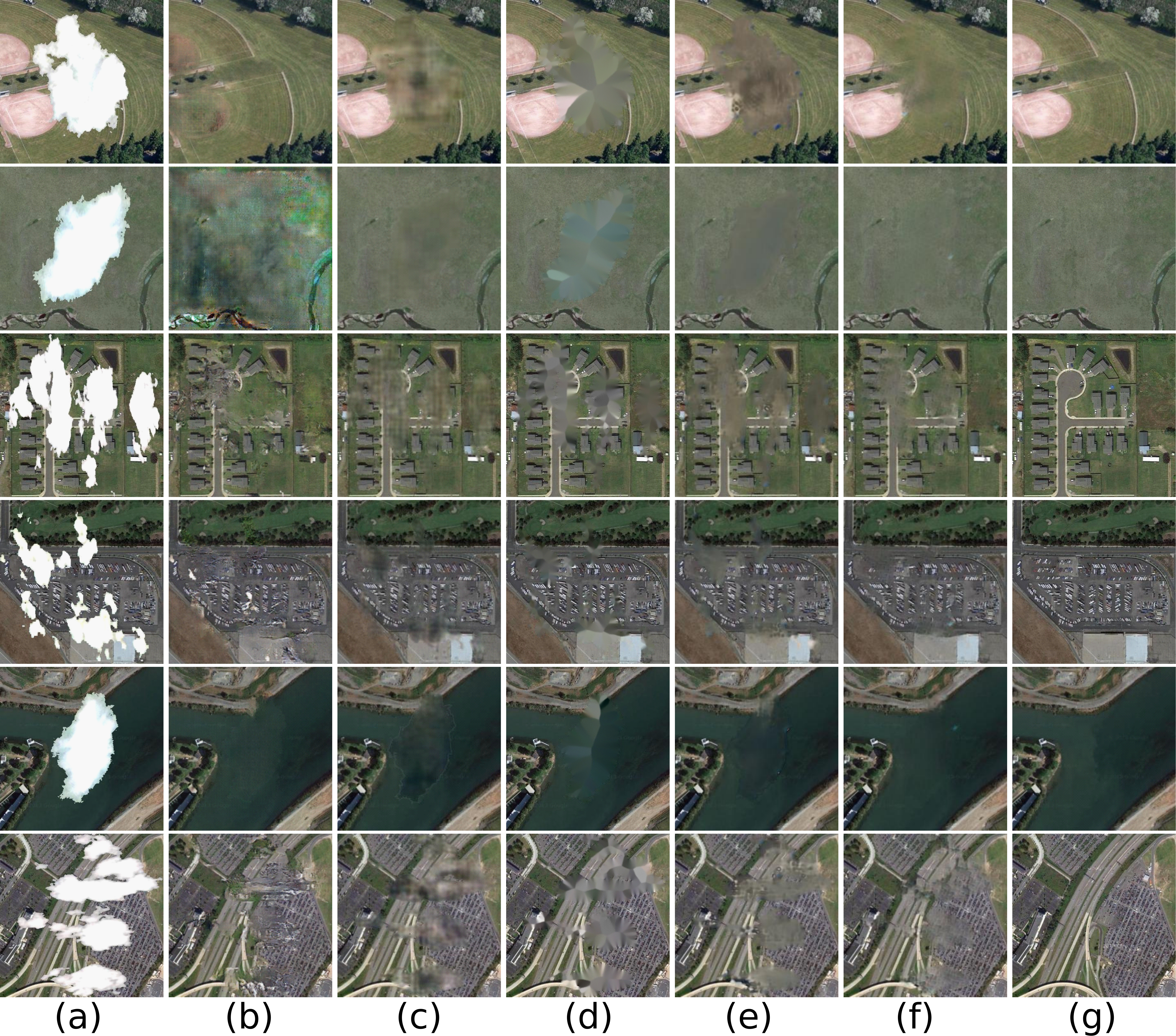}
\par\end{centering}
\caption{Qualitative comparisons on the RSSCN7 dataset: (a) Input images, (b-f) image inpainting results by  CSA~\cite{Liu_2019_CSA}, RFR~\cite{li2020recurrent}, 
TELEA~\cite{telea2004image},
MISF~\cite{li2022misf} and the Proposed method, (g) ground truth.} 
\label{fig:RSSCN7-demo}	
\end{figure}

\subsection{MaskMix based Data Augmentation}\label{Sec:MaskMix}
 
%--------------------------------------------------------------------
Since the geoscience data is always limited, not as large as the natural image dataset like ImageNet, we propose a novel data augmentation method for training image inpainting models to make better use of limited geoscience training data. 
Most of existing inpainting methods take the fixed masks, so some occlusion patterns are never processed when training the inpainting model, which limits the generalization capability of models to handle different occlusion scenarios. 
To address this issue, we propose to augment seed masks by conducting a series of transformations on the masks, which might happen in the real world, and mix them to simulate more complex scenarios.
Specifically, given a seed mask $M_s$ (\textit{e.g.}, the cloud in a satellite image), we transform it via different augmentation operations (\textit{e.g.}, translation, shearing, and rotation) and obtain multiple augmented masks. Intuitively, these processes might  simulate the cloud moving and reshaping in the real world. Finally, we mix all augmented masks to a single mask to obtain a complex occlusion pattern. Based on this idea, inspired by AugMix~\cite{hendrycks2019augmix}, we propose to augment the diversity of seed masks and increase robustness to unforeseen 
occlusion scenarios by a MaskMix based data augmentation. As shown in Fig.~\ref{fig:MaskMix}, we set three random  operations of ``translate", ``shear" and ``rotate" in parallel three branches to generate several random masks, then the MaskMix based data augmentation is computed by    
\begin{equation}
M = \Phi\{w_4 * M_s + \sum_{i=1}^{3} w_i * A^3_i(A^2_i(A^1_i(M_s))) \}, 
\label{eq:maskmix}
\end{equation}
where $M_s$ is the binary seed mask, $A^j_i$ is the randomly sampled \hm{augmentation} operations of ``translate", ``shear" and ``rotate" in the $i$-th row and $j$-th column as shown in Fig.~\ref{fig:MaskMix}, $w_i$ is the random sample mixing weight in the $i$-th row, $\Phi\{\cdot\}$ indicates the thresholding operation, and $M$ is the final augmented  binary mask by the MaskMix. $M$ is fed to train our proposed inpainting framework instead of the seed mask $M_s$. \hm{Different with the AugMix~\cite{hendrycks2019augmix} which augments the original image, the proposed MaskMix aims to augment the binary mask for the image inpainting problem.} Please note that the proposed MaskMix based data augmentation is only applied in the model training, not in the model testing.

\section{Experiments}~\label{Sec:Experiments}
% Overall, describe that we studied two Geoscience tasks in this paper: 1. RS scene recognition, 2. Cross-view geolocation 
In this section, we will evaluate the proposed method on \hm{three} widely-used geoscience tasks: \hm{remote sensing (RS) scene recognition, cross-view geolocation, and semantic segmentation}. The RS scene recognition task is to recognize an aerial-view satellite image into predefined classes~\cite{zou2015deep}, similar to the image classification problem. The cross-view geolocation task is to localize the spot by matching an given street-view image to the corresponding aerial-view satellite or UAV image in a gallery~\cite{wang2021each}, similar to the image retrieval problem. \hm{The remote sensing semantic segmentation task is to identify the land-cover or land-use category of each  part of the remote sensing High Spatial Resolution (HSR) image~\cite{wang2021loveda}.}

% add a picture to show the two datasets. % describe the task definition, and RSSCN7 dataset
\subsection{Dataset for RS Scene Recognition}\label{dataset:Rscene}
The dataset for the RS scene recognition task is the public aerial-view Google Earth satellite images, i.e., RSSCN7~\cite{zou2015deep} dataset. The RSSCN7 dataset contains satellite images acquired from Google Earth, which is originally collected for remote sensing scene classification. We conduct image synthesis on RSSCN7 to make it capable of the image inpainting task. It has seven classes: grassland, farmland, industrial and commercial regions, river and lake, forest field, residential region, and parking lot. Each class has 400 images, so there are total 2,800 images in the RSSCN7 dataset. 50\% is used for the network training, and another 50\% is for network testing in the RSSCN7 dataset. We first extract some thick/nontransparent clouds as 28 anchor masks from some real cloudy satellite images~\cite{tan2016cloud,cheng2017remote}. For each image of RSSCN7 dataset, we randomly pick one mask, and randomly rotate, translate, resize the mask and overlap it to the original image, and we make a constraint that the area ratio of the added occlusion over the whole image is between 15\% and 60\%. The  inpainting examples of occluded RSSCN7 dataset are shown in Fig.~\ref{fig:dataset}. 

\subsection{Dataset for Cross-view Geolocation}\label{dataset:geolocation}
% describe the task definition, and * dataset 
The public dataset used for the cross-view geolocation task is the CVUSA~\cite{zhai2017predicting} dataset. It includes the geoscience data collected from the ground view (street view) and satellite view. In CVUSA dataset, all the ground-view images are panoramic images downloaded from Google Street View, while the corresponding satellite-view images are collected from Microsoft Bing Maps. There are 35,532 ground-and-satellite image pairs for training and 8,884 image pairs for testing. The seed masks used to simulate occlusions are downloaded from the public image inpainting masks~\cite{liu2018image} and added to the ground-view images only. We use the seed masks with 10-20\% area ratio in the experiment. In this task, we assume that the satellite-view images are clean without occlusions. The examples of occluded CVUSA dataset for image inpainting are shown in Fig.~\ref{fig:dataset}. 

\subsection{Dataset for Semantic Segmentation}\label{dataset:geolocation}

\hm{The dataset used for semantic segmentation task is the public LoveDA~\cite{wang2021loveda} dataset. This dataset consists of rural and urban images that are obtained from Google Earth platform, along with their pixel-level labels. The LoveDA dataset contains 5,987 High Spatial Resolution (HSR) images in total, with image resolution of $1024\times1024$ pixels, whose spatial resolution is 0.3 m. In this experiment, we follow the default setting of  LoveDA~\cite{wang2021loveda} to have 4,191 images for training and the other 1,796 images for testing. As for the occlusion simulation, we employ the same strategy as that in cross-view geolocation task to generate occluded images and masks.}

% \subsection{Experimental setting, comparison methods and evaluation metrics}\label{exp:setting}
\subsection{Experimental Setups}\label{exp:setting}
 
{\bf Implementation details:} In the RS scene recognition experiment, the VGG16 based image classification network~\cite{simonyan2014very} is used as the geoscience task network. In the cross-view geolocation experiment, the Local Pattern Network (LPN)~\cite{wang2021each} is used as the geoscience task network. \hm{In the semantic segmentation experiment, HRNet~\cite{wang2020deep} is used as the geoscience task network.} The task networks are trained independently on the clean images without occlusions, and then they are fixed for task performance evaluation. We denote ``Clean Testing" as testing clean images without occlusions on the fixed geoscience task  network, and denote ``Occluded Testing" as testing the occluded images on the fixed geoscience task network. During the initialization of the proposed network, the first three layers of the encoders are initialized from the ImageNet pre-trained ResNet-34 \cite{he2016deep} model, and the other layers are randomly initialized. We set the loss balance weight $\lambda$ as 5 for RS scene recognition and semantic segmentation, 1.2 for cross-view geolocation. During the network training, each image is resized to $256\times256$ \hm{for RS scene recognition and cross-view gelocation task, $512\times512$ for semantic segmentation task.}  We use the PyTorch framework to implement the proposed network and all the experiments are run with a NVIDIA RTX 3090 GPU card. For more details, we will publicize the code after paper acceptance.   

% this is the setting of the RSSCN7 
%To train the proposed network, we set the learning rate as $2\mathrm{e}{-4}$, batch size as 8 and train the network until convergence. We apply the ``StepLR" strategy to decay the learning rate by 0.9 every 40 epochs. 
\begin{table}[t]
% by Yu 11/11/2020: directly use the predicted map to do the evaluation, no dot product
	\caption{\label{tab:classificaton results}	
	\hm{Quantitative experimental results on the RSSCN7 dataset for RS scene recognition using fixed VGG16~\cite{simonyan2014very} (pretrained on clean images) as the geoscience task  network.}}
	\small
	\begin{centering}
			\begin{tabular}{c|c|c|c}
			\hline 
			\hline 
			Methods & PSNR    & SSIM     &  Accuracy(\%)    
			\tabularnewline
			\hline 
			\hline 	
			Clean Testing~\cite{simonyan2014very} & - & - & 93.57 \tabularnewline
			\hline
			\hline 
			Occluded Testing~\cite{simonyan2014very} & 12.03 & 0.74 & 71.14 \tabularnewline
			\hline
			TELEA~\cite{telea2004image} & 25.20 & 0.78 & 76.86
			\tabularnewline 
			\hline
			Pix2Pix~\cite{isola2017image} & 24.18 & 0.73 & 83.14 
			\tabularnewline 
			\hline
		  SPL~\cite{zhang2021context}  & 25.43 & 0.829 & 84.21
	     \tabularnewline
	       \hline
		    GMCNN~\cite{wang2018image} & 25.18 & 0.78 & 81.07   \tabularnewline 
		    \hline
		    CSA~\cite{Liu_2019_CSA}  & 26.61 & 0.82 & 86.79
		   \tabularnewline

		    \hline
		    RFR~\cite{li2020recurrent}  & 26.81 & 0.81 & 87.14 
		     \tabularnewline
		       \hline
		  
		   MISF~\cite{li2022misf}  & 27.44 & 0.83 & 87.92 
		     \tabularnewline
		     \hline
		    Proposed$_b$ & 27.85 & 0.85 & 87.78   \tabularnewline
  \hline
		    Proposed- & 27.74
 & 0.84 & 89.85   \tabularnewline
		    \hline
		    Proposed & \textbf{28.25} & \textbf{0.86} & \textbf{90.21}   \tabularnewline
			\hline
			\hline
		\end{tabular}
		\par\end{centering}
\end{table}

{\bf Baselines:} We compare the proposed method with several state-of-art image inpainting or image transfer methods. These methods are RFR~\cite{li2020recurrent}, CSA~\cite{Liu_2019_CSA}, GMCNN~\cite{wang2018image}, TELEA~\cite{telea2004image}, Pix2Pix~\cite{isola2017image}, \hm{SPL~\cite{zhang2021context}, MISF~\cite{li2022misf}}. These comparison methods are carefully trained on the related geoscience datasets until convergence. We use \textit{Proposed$_b$} to represent our proposed baseline method (without Geoscience Task Network and MaskMix), \textit{Proposed-} as our proposed method (without MaskMix), \textit{Proposed} as our full proposed method. 

{\bf Metrics:} The evaluation metrics are two folds. One side is for the image quality evaluation using  structural similarity index (SSIM)~\cite{wang2004image} and peak signal-to-noise ratio (PSNR), following most image inpainting researches. The other side is for the geoscience task-related performance evaluation. In the RS scene recognition task, the overall classification accuracy (\%) on the whole testing set is used, following~\cite{zou2015deep}. In the cross-view geolocation task, we use Recall@K (R@K) and the average precision (AP) to evaluate the performance following~\cite{wang2021each}, where R@K represents the proportion of correctly matched images in the top-K of the ranking list. AP calculates the area under the Precision-Recall curve, which shows the precision and recall rate of the retrieval performance. \hm{For the semantic segmentation task, mean  intersection over union (mIoU) metric  is deployed to evaluate the performance.}  Higher value indicates the better performance for each metric.

\begin{table*}[htbp]
% by Yu 11/11/2020: directly use the predicted map to do the evaluation, no dot product
	\caption{\label{tab:geolocation_results}
	\hm{Quantitative experimental results on the CVUSA dataset for Cross-view geolocation using fixed LPN~\cite{wang2021each} (pretrained on clean images) as the geoscience task network.}}
	\small
	\begin{centering}
			\begin{tabular}{c|c|c|c|c|c|c|c}
			\hline 
			\hline 
			Methods & Recall@1    & Recall@5     & Recall@10  & Recall@top1\%  & AP  & PSNR & SSIM 
			\tabularnewline
			\hline 
			\hline 	
			Clean Testing~\cite{wang2021each} & 82.34  & 93.14   & 95.28  & 98.83 & 84.77 & - &  -
			\tabularnewline
			\hline
			\hline
            Occluded Testing~\cite{wang2021each} & 24.55  & 38.89   & 45.29 & 68.18  & 28.12 & 11.87 &  0.70
            \tabularnewline 
            \hline
			TELEA~\cite{telea2004image} & 43.85  & 61.85  & 68.65 & 86.44 & 48.10 & 24.17 &  0.80
			\tabularnewline 
			\hline
			Pix2Pix~\cite{isola2017image} & 40.05  & 57.85 & 64.40 & 83.53 & 44.32 & 23.76 & 0.76 
			\tabularnewline 
			\hline
	    SPL~\cite{zhang2021context}  & 52.62 & 70.33 & 75.98 & 90.03 & 56.73 & 24.74 & 0.85
	     \tabularnewline
	       \hline
		    GMCNN~\cite{wang2018image} & 46.25 & 63.81 & 70.14 & 86.44 & 50.40 & 26.23 & 0.84   \tabularnewline 
		    \hline
 		    CSA~\cite{Liu_2019_CSA}  & 52.47  & 70.66  & 76.60 & 91.32 & 56.73 & 26.26 & 0.89
		   \tabularnewline

		    \hline
		    RFR~\cite{li2020recurrent}  & 47.55 & 66.05  & 72.29 & 87.83  & 51.88 &  26.84 & 0.88
		     \tabularnewline
		     \hline
		 
	         MISF~\cite{li2022misf}  & 69.64 & 85.92 & 89.94  & 97.07  & 73.33 & 26.73 & 0.86
		     \tabularnewline
		     \hline
		    Proposed$_{b}$ &  68.20  & 84.68   & 88.87 & 96.74    & 71.93 & \textbf{28.47}& \textbf{0.92} \tabularnewline
  \hline
		    Proposed-   & 75.46 & 89.46  & 92.38 & 97.75  & 78.62 &  28.34 & 0.91 \tabularnewline
		    \hline
		    Proposed   & \textbf{75.69} & \textbf{89.55}   & \textbf{92.55} & \textbf{97.76}  & \textbf{78.84} &  28.37 & 0.91  \tabularnewline
			\hline
			\hline
		\end{tabular}
		\par\end{centering}
\end{table*}

% a table (image quality, task related) and a figure 

% Proposed_{b}_{b}: baseline  
% Proposed: baseline + task loss   
% Proposed+: baseline + task loss+ MaskMix

\begin{figure*}[htbp]
\begin{centering}
\includegraphics[width=0.95\textwidth]{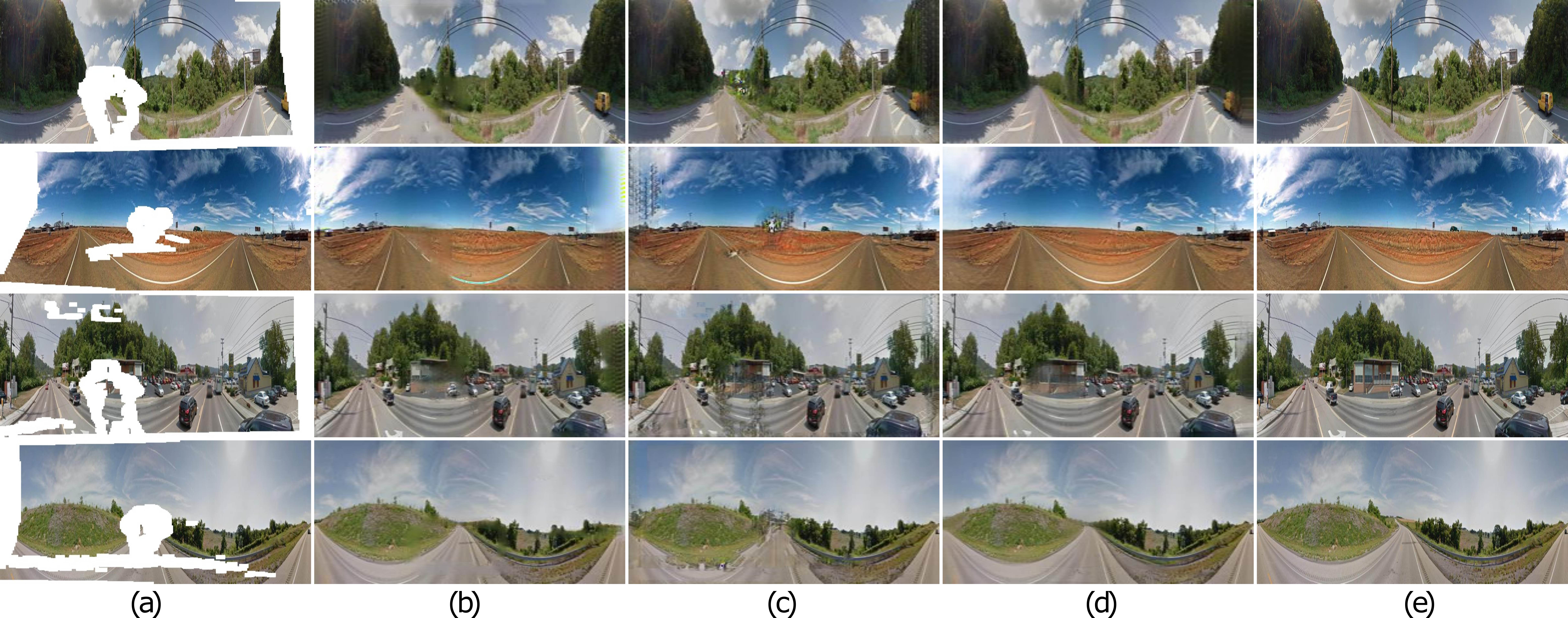}
\par\end{centering}
\caption{Qualitative comparisons on the CVUSA dataset: (a) Input ground/street-view images, (b-d) image inpainting results by  CSA~\cite{Liu_2019_CSA}, RFR~\cite{li2020recurrent}, and the Proposed method, (e) ground truth.} 
\label{fig:CVUSA-demo}	
\end{figure*}

\subsection{RS Scene Recognition Results}
\label{result:RSscene}

Table~\ref{tab:classificaton results}  shows the quantitative performance  on the RSSCN7 dataset. When there are no occlusions, the geoscience task network gets 93.57\% overall accuracy, seeing ``Clean Testing". If we feed the occluded images into the fixed geoscience task network pre-trained on clean images, the overall accuracy for ``Occluded Testing" is only 71.14\%, since it is more difficult for the fixed geoscience task sub-network to make accurate recognition when there are occlusions in image. With each of the image inpainting methods, the recognition accuracy is improved, this improvement indicates that image inpainting could help the geoscience related task. Among the comparison methods, \hm{the MISF~\cite{li2022misf} method got the advanced performance of PSNR 27.44, SSIM 0.83, Accuracy 87.92\%.} However, the Proposed method obtained the best performance of PSNR 28.25, SSIM 0.86, Accuracy 90.21\%. This phenomenon demonstrates that the proposed method could not only achieve the best task-related performance, but also obtain advanced reconstructed image quality. The qualitative results of image inpainting \hm{for RS scene recognition} are shown in Fig.~\ref{fig:RSSCN7-demo}.       

Comparing  the Proposed$_b$,  Proposed-, and  Proposed versions of our method, we can see that 1) the proposed coarse-to-fine baseline method is with good-quality reconstruction and reasonable task-related performance; 2) adding the geoscience task network into the proposed  framework is helpful in the RS scene recognition task, with only slightly decrease in reconstructed image quality; 3) further including MaskMix in the proposed framework could continue to improve the task-related performance and reconstructed image quality.

% \subsection{Experimental results for Cross-view geolocation}\label{result:geolocation}
\subsection{Cross-view Geolocation Results }\label{result:geolocation}
% a table and a figure 
Table~\ref{tab:geolocation_results}  shows the quantitative results on the CVUSA dataset for the cross-view geolocation task. With the fixed LPN~\cite{wang2021each} pretrained on clean images as the geoscience task network, clean images could obtain 84.77\% AP while occlusions will reduce it to 28.12\%. It demonstrates that occlusion leads to the significant challenge to the cross-view geolocation problem. By each  image inpainting method, the task-related performance could be increased. Among them, the Proposed method gets the best AP as 78.84\%, much larger than other image inpainting methods. This is because that the Proposed method is designed to improve the geoscience task-related performance without changing the fixed geoscience task network. Reconsidering 1) the large increase from 28.12\% AP to our 78.84\% improvement and 2) no need to change the  fixed geoscience task network pretrained on clean images, our proposed method is quite promising to process and understand the occluded geoscience images.  Specifically, the Proposed$_b$ already obtains better performance than other comparison methods on both image quality evaluation metrics and geoscience task-related evaluation metrics. When combined with geoscience task network (Proposed-) and MaskMix (Proposed), the task-related performance, R@K and AP, could be further improved, which is consistent with the experimental results for RS scene recognition.

\hm{It is worth mentioning that the image quality (PSNR, SSIM) of final Proposed method is relatively high  but not the best, compared to Proposed$_{b}$. \textit{It also verifies that better image quality not always lead to higher geoscience task performance}. Our goal is applying inpainting to reach much better geoscience task performance along with relatively good image quality, without changing the fixed geoscience task network pretrained on clean images. Some previous computer vision research also shows that only enhancing the image reconstruction quality does not necessarily improve the next computer vision task performance. As shown in~\cite{hnewa2020object}, deraining even decreases object detection accuracy on the rainy images. As studied in~\cite{pei2019effects}, removing haze cannot largely improve its classification performance.} The qualitative results of image inpainting \hm{for cross-view geolocation} are shown in Fig.~\ref{fig:CVUSA-demo}.

\subsection{\hm{RS Semantic Segmentation Results} }\label{result:geolocation}
\begin{figure}[htbp]
\begin{centering}
\includegraphics[width=1\columnwidth]{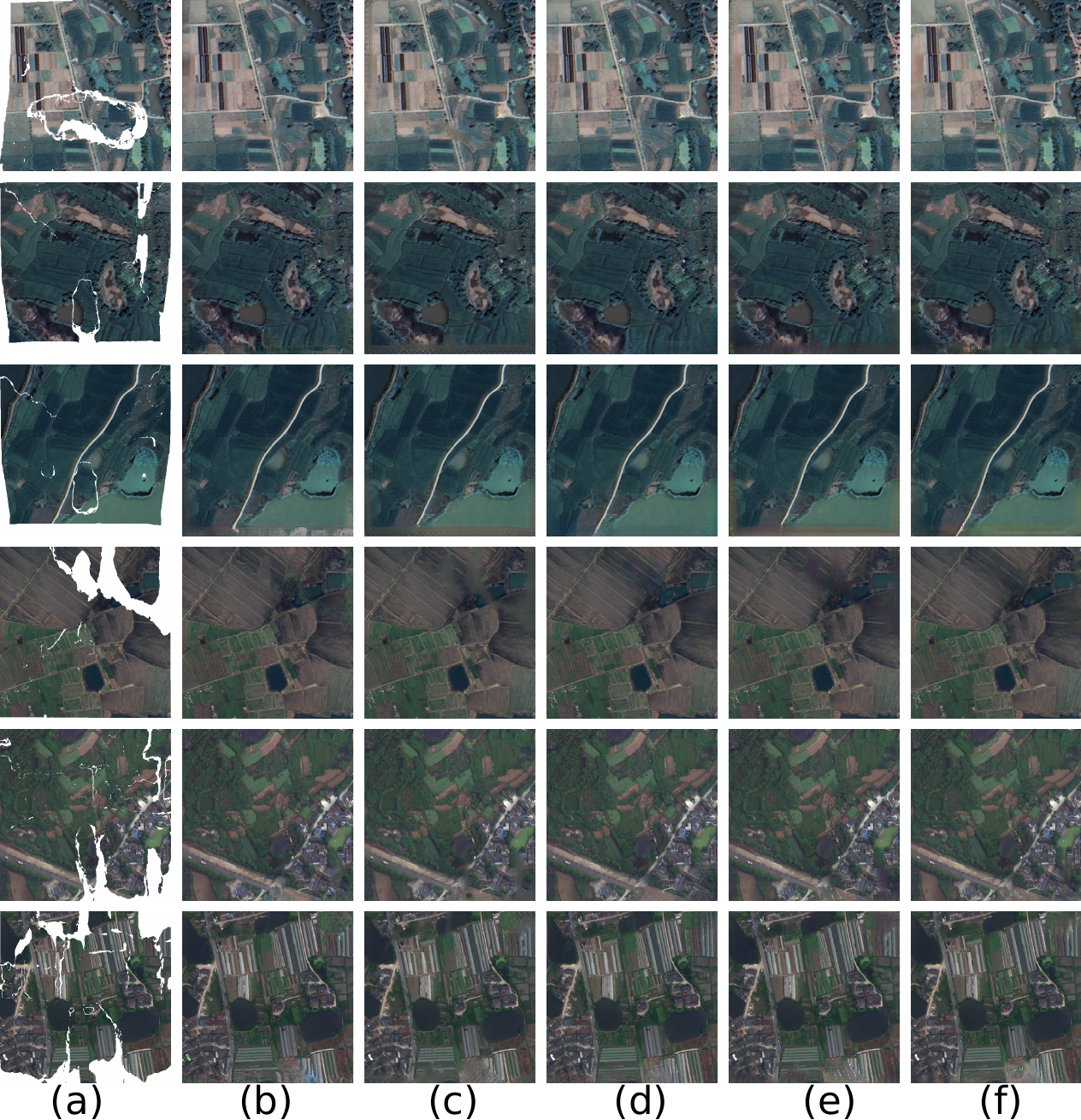}
\par\end{centering}
\caption{\hm{Qualitative comparisons on the LOVEDA dataset: (a) Input images, (b-e) image inpainting results by  CSA~\cite{Liu_2019_CSA}, RFR~\cite{li2020recurrent},  MISF~\cite{li2022misf} and the Proposed method, (f) ground truth.}} 
\label{fig:LOVEDA-demo}	
\end{figure}

\hm{Table \ref{tab:segmentation results} shows the quantitative performance on the LOVEDA dataset. When there are no occlusions, the segmentation task network gets 0.4403 mIoU. If the occluded images are fed into the semantic segmentation task network pre-trained on clean images of the LOVEDA dataset, the mIoU for “Occluded Testing” is only 0.3554. The mIoU performance could be improved by different inpainting methods. These improvements demonstrate that inpainting methods help to reduce the impact of occluded regions on geoscience task. Among the comparison methods, the SPL method obtains the advanced performance of PSNR 30.90, SSIM 0.93, mIoU 0.4176. However, the proposed method achieves the best mIoU 0.4329, with comparable PSNR and SSIM score of 29.2 and 0.92. This phenomenon indicates that although the proposed method did not reach the highest image reconstruction quality in PSNR and SSIM, it could obtain the best task-related performance. The qualitative results of image inpainting for semantic segmentation are shown in Fig.~\ref{fig:LOVEDA-demo}. The proposed method could generate comparable image reconstruction quality as other methods while obtain the best task-related results at the same time.}

\begin{figure}[htbp]
\begin{centering}
\includegraphics[width=0.95\columnwidth]{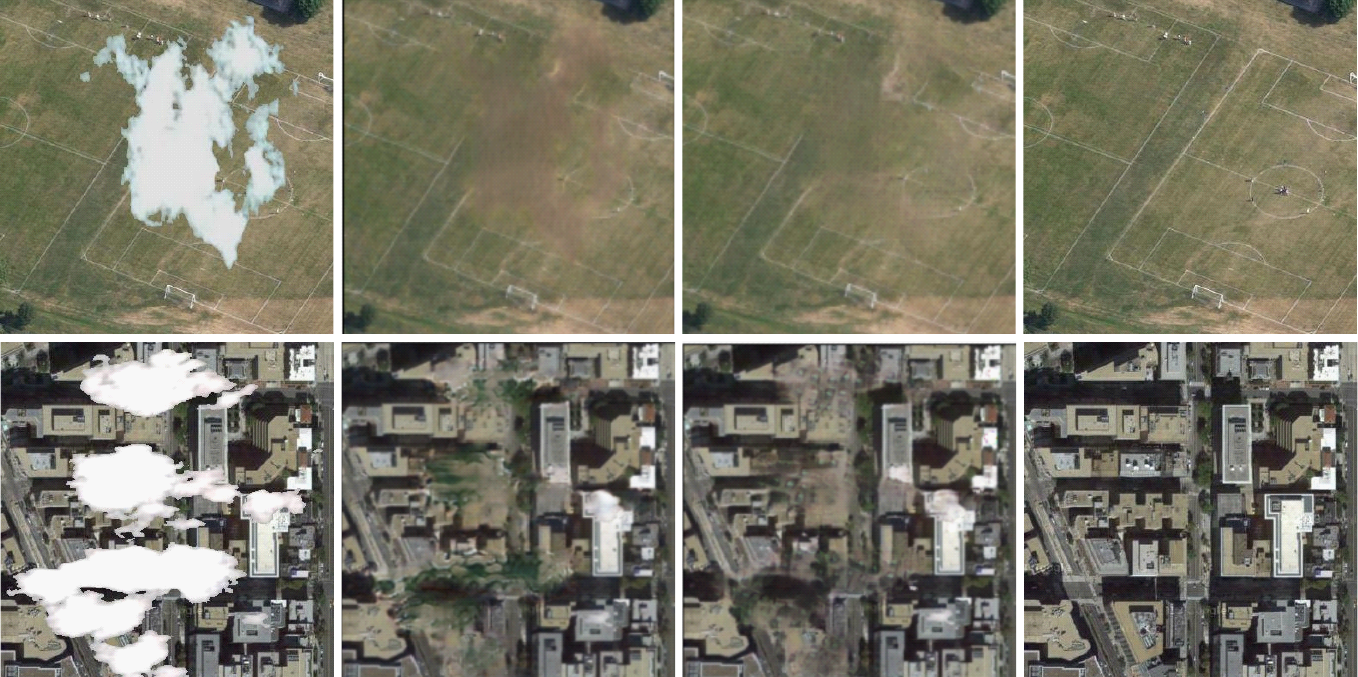}
\par\end{centering}
\caption{Illustration of the coarse and refined reconstruction maps by the proposed image inpainting method using Proposed$_b$ as example. From left to right: input satellite image with occlusions, coarse reconstruction map, refined reconstruction map, and ground truth.}
\label{fig:coarse_refined_example}	
\end{figure}

\begin{table}[t]
	\caption{\label{tab:segmentation results}	
	\hm{Quantitative experimental results on the LOVEDA dataset for RS semantic segmentation  using fixed HRNet~\cite{wang2020deep}  (pretrained on clean images) as the geoscience task  network.}}
	\small
	\begin{centering}
			\begin{tabular}{c|c|c|c}
			\hline 
			\hline 
			Methods & PSNR    & SSIM     &  mIoU    
			\tabularnewline
			\hline 
			\hline 	
			Clean Testing~\cite{wang2020deep} & - & - & 0.4403 \tabularnewline
			\hline
			\hline 
			Occluded Testing~\cite{wang2020deep} & 12.13 & 0.85 & 0.3554 \tabularnewline
			\hline
	
		    CSA~\cite{Liu_2019_CSA}  & 28.76 & 0.92 & 0.4231
		   \tabularnewline
		    \hline
		    RFR~\cite{li2020recurrent}  & 30.15 & 0.92 & 0.4132 
		     \tabularnewline
		       \hline
		   MISF~\cite{li2022misf}  & 30.76 & 0.92 &  0.4271 
		     \tabularnewline
		     \hline
		       SPL~\cite{zhang2021context}  & \textbf{30.90} & \textbf{0.93} & 0.4176
	     \tabularnewline
	       \hline
		    Proposed$_b$ & 28.45 & 0.92 & 0.4283 \tabularnewline
  \hline
		    Proposed- & 29.21
 & 0.92 & 0.4320  \tabularnewline
		    \hline
		    Proposed & 29.20 & 0.92 & \textbf{0.4329}   \tabularnewline
			\hline
			\hline
		\end{tabular}
		\par\end{centering}
\end{table}

\subsection{Effectiveness of Coarse~\&~Refined Inpaintings}

\hm{In this section, we study the effectiveness of coarse and refined inpaintings respectively}. Taking the Proposed$_{b}$ method on RSSCN7 dataset as an example, the coarse reconstruction gets PSNR 27.10 and SSIM 0.84, while the coarse-to-fine reconstruction obtains better PSNR 27.85 and SSIM 0.85. Therefore, the coarse-to-fine learning performs better than the coarse network learning only. As shown in Fig.~\ref{fig:coarse_refined_example}, compared to the coarse reconstruction map, the refined reconstruction map looks more smooth with less defects in our experiment.

\begin{figure}[htbp]
\begin{centering}
\includegraphics[width=0.95\columnwidth]{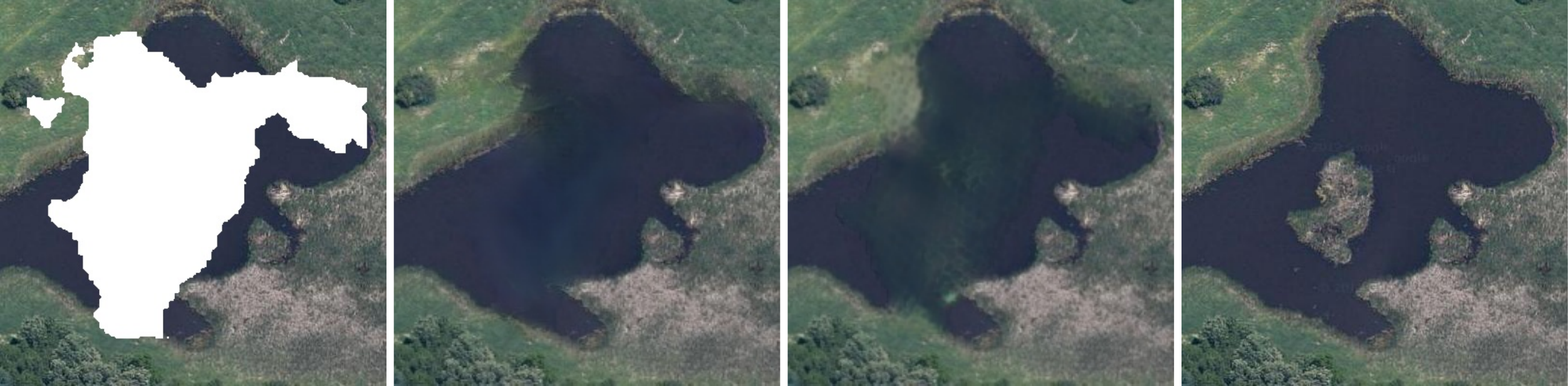}
\par\end{centering}
\caption{\hm{Illustration of the failure case by the proposed image inpainting method. From left to right: input satellite image with occlusion, result by MISF~\cite{li2022misf}, result by proposed method, and ground truth.}}
\label{fig:FailureCase}	
\end{figure}

\subsection{Running Time \& Failure Case}
During the testing stage, our running time per $512 \times 512$ RGB color image inpainting is 0.1s, and our running time  per $256 \times 256$ RGB color image inpainting is 0.033s. This demonstrates the proposed network is efficient for geoscience image inpainting task and could be used as an image pre-processing step.

\hm{As shown in Fig.~\ref{fig:FailureCase}, if an object (like an island) of geoscience image is fully covered by occlusion, the proposed method could not well reconstruct it because the proposed method does not have enough prior knowledge of the geoscience context. This disadvantage of the proposed method could be overcame if feeding with more context images, \textit{e.g.}, other geoscience images without occlusion of the same location in different time. This is different with the common image impainting problem with rich prior knowledge, \textit{e.g.}, the face image impainting.}
%------------------------------------------------------------------------- 
\section{Conclusions}\label{Sec:Conclusions}
In this paper, we studied the research problem of task-driven image inpainting for geoscience images. Instead of only going for a better image quality after reconstruction, our objective is to largely improve the geoscience task performance with relatively high image quality, without changing the fixed/deployed geoscience task network pre-trained on clean images. We proposed a coarse-to-fine task-driven learning based deep CNN model with MaskMix based data augmentation for this problem. \hm{The experimental results on RS scene recognition, cross-view geolocation, and semantic segmentation tasks show the effectiveness and accuracy of the proposed method.}

\ifCLASSOPTIONcaptionsoff
  \newpage
\fi

% \end{thebibliography}
\bibliographystyle{IEEEtran}
\bibliography{huiming}

\end{document}